\pdfoutput=1

\documentclass[11pt]{article}

\usepackage{ACL2023}

\usepackage{times}
\usepackage{latexsym}
\usepackage{graphicx}
\usepackage{multirow}
\usepackage{enumitem}
\setitemize{fullwidth}

\usepackage[T1]{fontenc}

\usepackage[utf8]{inputenc}

\usepackage{microtype}

\usepackage{inconsolata}

\title{CiteCaseLAW: Citation Worthiness Detection in Caselaw for Legal Assistive Writing}

\author{Mann Khatri \\ Indraprastha Institute \\ of Information \\ Technology, Delhi \And
        Pritish Wadhwa \\ Indraprastha Institute \\ of Information \\ Technology, Delhi
        \And
        Gitansh Satija \\ Indraprastha Institute \\ of Information \\ Technology, Delhi \And
        Reshma Sheik \\ National Institute \\ of Technology, \\ Tiruchirappalli 
        \AND
        Yaman Kumar \\ Adobe MDSR  \And
        Rajiv Ratn Shah \\ Indraprastha Institute \\ of Information \\ Technology, Delhi
        \And
        Ponnurangam Kumaraguru \\ International Institute \\ of Information \\ Technology, Hyderabad
        }

\definecolor{DodgerBlue3}{rgb}{0.12, 0.56, 1.0}
\definecolor{Aquamarine}{rgb}{0.5, 1.0, 0.83}

\begin{document}
\maketitle
\begin{abstract}

In legal document writing, one of the key elements is properly citing the case laws and other sources to substantiate claims and arguments. Understanding the legal domain and identifying appropriate citation context or cite-worthy sentences are challenging tasks that demand expensive manual annotation. The presence of jargon, language semantics, and high domain specificity makes legal language complex, making any associated legal task hard for automation. The current work focuses on the problem of citation-worthiness identification. It is designed as the initial step in today's citation recommendation systems to lighten the burden of extracting an adequate set of citation contexts. To accomplish this, we introduce a labeled dataset of 178M sentences for citation-worthiness detection in the legal domain from the Caselaw Access Project (CAP). The performance of various deep learning models was examined on this novel dataset. The domain-specific pre-trained model tends to outperform other models, with an 88\% F1-score for the citation-worthiness detection task. 


\end{abstract}

\section{Introduction}
Legal artificial intelligence has advanced rapidly with natural language processing (NLP) and deep learning in the last few decades. Due to the highly domain-specific and varied nature of legal language and the need for significant annotated data sources, developing effective NLP systems from legal text is challenging. In addition to the inherent difficulties with legal text, such as its lengthy content, distinctive internal structure, and domain-specific terminologies, the processing of external legal citations is the most significant and noticeable issue \cite{martin2012basic}.

Legal citations are essential to the functioning of a case-law-based legal system. A necessary part of writing any legal document is appropriately citing sources. The references cited by legal professionals in legal decisions show how those cases are connected to the current case. A citation-worthiness identification procedure, often referred to as citation context detection, is designed as the first stage in present citation recommendation systems. For the goal of comprehending the law and applying it correctly to new cases, the treatment given to the cited cases is a significant element. Due to the extremely high volume of court decisions rendered daily,  legal professionals require assistance in discovering relevant citation-worthy sentences. Finding citation-worthy sentences or statements that contain a reference to an outside source is the problem of citation-worthiness detection \cite{Citationworthinessinscientic}. An automated system should be able to determine whether a citation is necessary for a sentence given from a legal source without using any specific knowledge base. The sentence considered a potential citation candidate might cite one or more sources. We aim to identify which informational components are necessary for the sentence to be marked as citation-worthy. The citation worthiness task is a binary classification task used to place a sentence into the ``cite'' or ``not cite'' category.
Table~\ref{Ex} provides examples for each binary label, highlighting the differences between ``cite'' and ``not cite'' sentences in legal decisions.

\begin{table*}[h]
    \centering
   \begin{tabular}{@{}p{14cm}@{}p{0.5cm}p{1cm}}
    \hline\vspace{1pt}
\textbf{Example}& &\vspace{1pt}\textbf{Label}\\
        \hline \vspace{1pt}
\textcolor{teal}{This statute applies to alimony obligations created by verdict.} \textcolor{cyan}{See Allen v. Allen, 265 Ga. 53 (1) (452 SE2d 767) (1995)} &&\vspace{1pt} \textbf{cite} \\
\hline \vspace{1pt}
\textcolor{teal}{However, while an exemption should be strictly construed, the construction must still be reasonable. } \textcolor{cyan}{Trustees of Ind. Univ. v. Town of Rhine, 170 Wis. 2d 293, 299, 488 N.W.2d 128 (Ct. App. 1992).}&&\vspace{1pt}\textbf{cite}\\
\hline \vspace{1pt}
This leaves absolutely indefinite and uncertain what the plaintiff was to receive.&&\vspace{1pt}\textbf{not\_cite}\\
\hline \vspace{1pt}
The appellant then was granted the right and did file amendments to its assignments of error.&&\vspace{1pt}\textbf{not\_cite}\\
\hline
    \end{tabular}
    \caption{Example sentences for ``cite'' and ``not\_cite'' labels from CiteCaseLAW. The citation-worthy sentence appears in green, and the sentence marked in blue seems to confirm that claim.}
    \label{Ex}
\end{table*}

The task is beneficial in reducing the legal professional's effort in drafting legal decisions. If a sentence is not identified as a citing sentence, it will not be examined in the remaining steps of the recommender system. The recommendation system must work more diligently when a large number of sentences are classified as citing sentences\cite{2022citationworthy}. A proper balance between citing and not-citing sentences would be of considerable interest. Thus the effectiveness of the subsequent stages heavily depends on the outcomes of the citation-worthiness identification task and helps in intelligent writing assistants. Two main obstacles to constructing a successful legal citation detection system are data volume and quality \cite{lourentzou2019data}. Building a large, high-quality labeled legal corpus is necessary for training deep learning models, but generating a manually annotated corpus is expensive. To the best of our knowledge, there is no sizeable dataset for citation-worthiness detection in the legal domain. 

Our main goal is to create a large dataset for the citation-worthiness detection task at the sentence level for the American legal domain. This task will act as a foundation for several applications requiring assistants for legal writing in the future. 

Creating a dataset for citation-worthiness detection involves extracting sentences from a legal document, labeling each sentence as to whether it contains a citation, and eliminating all citations. We then employed different machine learning and deep learning models to evaluate this dataset, selecting the optimal model  for downstream legal tasks. Thus we aim to answer the following research questions.\\
\textbf{RQ1:} How can a dataset for citation-worthiness detection for legal domain be automatically created with low noise without the support of domain-specific tokenizers/segmenters?\\
\textbf{RQ2:} What techniques are more reliable for identifying citation-worthiness sentences in the legal domain? \\
\textbf{RQ3:} As humans write cases, can we employ some strategy to remove subjectivity in writing?\\
\textbf{RQ4:} How do the models developed on the citation-worthiness dataset compare to the established baselines for other legal text classification tasks?\\
Following is a summary of the major contributions made in this research:
\begin{enumerate}[noitemsep]
\item We offer a novel dataset for citation worthiness detection task by extracting data from the Caselaw Access Project (CAP).\footnote{\url{https://case.law/}. CAP has two data sources, the Harvard Law School Collection, and the Fastcase collection. It uses Optical Character Recognition (OCR) for digitalizing the cases from paper.} Our corpus has 178M sentences for the citation-worthiness detection task and will be made publicly available after acceptance of the paper. Examples can be found here.\footnote{\url{https://anonymfile.com/VpWJq/examples100.jsonl}} 

\item We experimented with numerous state-of-the-art models to quantitatively assess them and maintain them as baselines for the task of citation worthiness detection. We also observed that the models that had been pre-trained in the legal domain performed better, demonstrating the usefulness of CiteCaseLAW for the task of citation-worthiness detection.

\item We also considered the removal of subjectivity induced in legal writing for detecting citation-worthy sentences. Further, we extend our experiment on how well this dataset can be adopted by a model fine-tuned on different legal benchmark datasets without negatively affecting its performance.
\end{enumerate}
The rest of this paper is organized as follows. The next section presents the existing approaches to citation-worthiness detection tasks. Following that, we discuss the annotation experiments for creating a corpus for the legal citation-worthiness detection task. It is followed by a thorough investigation of the baseline models developed for the corpus. Based on the assessment results, we present experiments with several downstream legal tasks with the best-performing model. Finally, we conclude this paper with limitations and future directions.
\section{Related Work}
In this section, we will cover the work done in the area of citation-worthiness on different datasets in the domain of science due to the lack of research in the legal domain. Next, we shall address how knowledge of these citation patterns can be extracted in legal language, featuring the challenges faced in Legal Citation for the worthiness detection task.
\subsection{Citation-Worthiness in Scientific Texts}
Citation worthiness is the task of detecting sentences that will need citation from an external source. One can cite some work to give credit or support the underlying argument. However, most of the efforts in citation-worthiness detection are in the scientific domain. \citeauthor{sugiyama2010identifying} created a dataset from the ACL Anthology Reference corpus (ACL-ARC, \cite{bird2008acl}), and using heuristics, they removed citation markers. Further, SVM with unigrams, bigrams, the presence of proper nouns, and the classification of previous and next sentences was used for citation-worthiness detection. Convolutional recurrent neural networks were used by \citeauthor{farber2018cite} on three datasets from ACL-ARC, arXiv CS \citeyearpar{farber2018high}, and Scholarly Dataset 2.\footnote{\url{http://www.comp.nus.edu.sg/~sugiyama/SchPaperRecData.html}} A similar strategy was applied by \citeauthor{bonab2018citation}. The context is not modeled in any of the aforementioned works. \cite{gosangi2021use} introduced the contextual dataset, ACL-cite, and they used BiLSTMs (Bidirectional Long Short-Term Memory) and transformer-based contextual word embeddings to build a context-aware model using context windows. On the citeWorth dataset, \cite{wright2021citeworth}  conducted an extensive study, including domain adaptation and transfer learning, developing a context-aware model. A BiLSTM-based architecture was developed by  \cite{zeng2020modeling}, which also demonstrated how context, and more specifically the context of the two adjacent sentences, may help to improve the prediction of citation quality. Later in \cite{2022citationworthy} focused on sentence-level citation worthiness identification as an important phase of citation recommendation systems. They conducted an in-depth section-wise analysis of the ACL-ARC dataset and offered a better model utilizing a syntax-based learning strategy to generate a low-dimensional representation of words intended to cover long-distance dependency. Furthermore, they used several down-sampling analyses to assess the model's performance to get balanced citation-worthiness identification data.

When extracting implicit citations for scientific articles, \cite{Jebari2018implicit}  employed an unsupervised method in which the sentences that follow an explicit citation are considered candidates for implicit citation.  They used the word embedding models Sentence2Vec and Topic2Vec to gauge how close the candidate sentence was to the cited work and labeled the most similar candidate sentence as the implicit citation. Another approach \cite{metadatacitation} used freely available metadata-based  parameters to extract the implicit citations. They used different machine learning classifier models, such as the Support Vector Machine (SVM), Kernel Logistic Regression (KLR), and Random Forest, to automatically classify citations into significant and non-significant categories. In \cite{Scicite2019} introduced a large dataset of citation intents and proposed a multitasking framework for predicting section headings and citation worthiness. Using this dataset \cite{ImpactCite2020}  proposed a new dataset eliminating duplicates and inconsistent labels to provide fair and meaningful results using cross-validation to overcome the limited number of examples for minority classes. The critical analysis of existing datasets was presented in \cite{CitationIntent} and proposed a text clustering-based mechanism to annotate the unlabeled dataset using the citation context.
\subsection{Legal Citation}
In \cite{sanchez2019sentence}, they looked at various methods for spotting sentence breaks in legal language. Due to its complexity of punctuation and syntax, legal literature poses difficulties for sentence tokenizers. Legal material is difficult for out-of-the-box algorithms to perform well on, which hinders future text analysis. Works of \citeauthor{sadvilkar-neumann-2020-pysbd} \citeyearpar{sadvilkar-neumann-2020-pysbd} showed that a general statistical sequence labeling model is capable of learning the definition more efficiently by creating a data set of 80 court decisions from four different domains and concluded that legal decisions are more challenging for existing sentence boundary detection systems than for non-legal texts and existing sentence boundary detection systems generally fails on legal tasks.

The full text of more than a million appeal decisions from 1999 to 2017 is available in the BVA corpus that \cite{huang2021context} use. A set of metadata derived from the Veterans Appeals Control and Locator System (VACOLS) is provided with each decision. This metadata includes fields like the decision date, diagnostic codes indicating the veteran's injuries, the case outcome, and a flag indicating whether the case was subsequently appealed. Each case also includes one or more ``problem codes," which are manually assigned by BVA lawyers. It then groups the main legal or factual issues highlighted (for example, ``entitlement to a funeral benefit"). Although our techniques can be applied to the entire corpus, this work concentrates on a subset of 3,24,309 cases that raise a single problem and have comprehensive metadata.
This dataset is a subset of Caselaw Access which we have taken and processed according to our task. They include the Vet. App. and F.3d reporters\footnote{A series of volumes known as "law reports" or "reporters" contains judicial judgments drawn from a variety of case law decided by courts. A list of reporters can be found at \url{https://api.case.law/v1/reporters/}}, which contain veterans’ law cases and cases from the Federal Courts of Appeal, as these account for the vast majority of cases cited in the corpus.

There is a complete lack of any dataset suitable for identifying citation-worthy sentences in the legal domain. We have addressed this problem by creating a sizeable dataset consisting of 178M clean and high-quality sentences.

\section{Dataset}
Answering \textit{RQ1} in this section. CAP is a repository of American legal cases from all the state, federal and territorial courts. All the legal cases are categorized into 61 different jurisdictions. Each legal case contains details about the presiding judge, reporter, court of jurisdiction, and cited cases. CAP also provides an OCR confidence score for all the cases. For our task, we extracted the dataset version 3 provided by the CAP. It was last updated on September 21, 2021.

\subsection{Data Visualization}
In this sub-section, we visualize randomly sampled court cases across different jurisdictions for conformity within them and also analyze the writing style of cases over centuries.

\begin{figure}[!ht]
  \centering
  \includegraphics[width=0.8\linewidth]{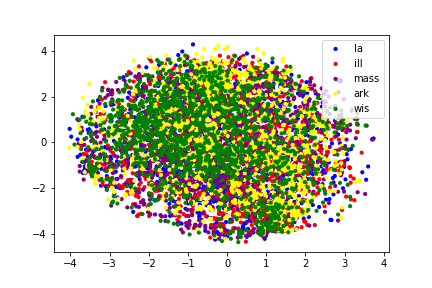}
  \caption{Visualizing LEGAL-BERT Embeddings for the dataset for five jurisdictions which shows that there are no observable differences between any two different jurisdiction. The jurisdictions chosen for this were \textit{`la`, `ill', `mass', `ark' and `wiss'}.}
  \label{fig1}
\end{figure}

\begin{figure}[!h]
  \centering
  \includegraphics[width=0.8\linewidth]{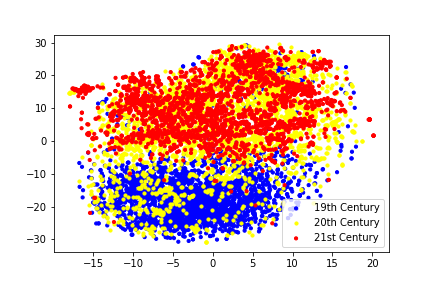}
  \caption{Visualizing LEGAL-BERT Embeddings for the 'ill' jusrisdiction over three centuries which shows the gradual shift in the writing style of the legal documents along the time period.}
  \label{fig3}
\end{figure}

We randomly sampled 1,500 cases from five randomly selected jurisdictions. Using LegalBert, we extract embeddings of each case and generated the TSNE plot of these embeddings. Figure \ref{fig1} shows the distribution of the data belonging to the individual jurisdictions implying that the results of the proceedings were independent of the jurisdictions.
Further, we plotted 2,500 cases randomly from every jurisdiction ranging from the 19\textsuperscript{th} to 21\textsuperscript{st} century. Figure~\ref{fig3} shows a toy example of \textit{`ill'} jurisdiction. These clusters indicate a gradual change in a court case's writing style, supporting our argument that there is subjectivity in writing the legal case. We address subjectivity later in the paper.

\subsection{Data Preprocessing}
In CAP, court cases are presented in different sections. The `Opinion' section contains the case transcript. If present, the excerpts from other legal cases and their citations were removed. These excerpts were irrelevant as they could interfere with the continuity of sentences. Footnotes, page numbers, and Non-ASCII tokens were fixed when they appeared between the texts. We created and applied regex patterns to remove such noise. 

To prepare our data preprocessing methodology, we thoroughly analyzed 500 documents across cases of different jurisdictions across centuries. Then, we examined the data cleaning outcomes using 500 additional documents chosen at random. All the 1000 documents were noise-free, and the desired textual data for each case was procured correctly after our preprocessing.

\subsection{Sentence Boundary Detection}
Extraction of the whole sentence from a legal document is challenging. In order to create a noise-free dataset, sentence boundaries must be detected accurately. We tried using popular sentence splitters, including SpaCy \cite{spacy2}, NLTK \cite{nltk}, and SegTok \footnote{https://github.com/fnl/segtok}. They are not designed to handle boundaries in the general English domain, making them weak segmenters. Work done by \citeauthor{sadvilkar-neumann-2020-pysbd} has a set of golden rules to detect boundaries, improving splitting on our dataset. OCR induces some errors, including incorrect capitalization of letters. We created a list of such tokens that hindered the sentence-splitting process and replaced them with suitable tokens. Examples are in the appendix section \ref{appendixsbd}.

Table \ref{tab:segtokexamples} of appendix section \ref{sectionG.3} shows the result of sentence splitters on Legal Text. We analyzed 30 documents containing about 1,800 sentences, and none of the traditional sentence splitters could classify more than 50\% correct sentences except for pySBD. We validated our sentence-splitting process by randomly sampling 50 documents containing around 2700 sentences in total and found that apart from 4 sentences, all others were correctly split.

\subsection{Citation Detection}

Citation list provided in the metadata (\ref{appendixmeta}) for each document. The list did not contain all the citations present and several citations were present as italicized text. For these few citations, we wrote a regex after analyzing documents from different time periods and courts. Some example citations are listed in the appendix section \ref{appendixsbd}.

We categorized the citation formats into two different types. First, was `versus' type citation, i.e., `Party A  vs.  Party B'. A regex was built to detect the span of such citations. The other type of citation represented the case id containing details of the reporter, court, year of the case, and such information. We developed a different rule-based approach to identify such types of citations. We extended our regex to capture any page number, volume number, in-line quotation, and other such data that occasionally followed the citation. We identify four types of sentences which are defined as follows and table \ref{tab:processedSentencesExamples} of appendix section \ref{sectionG.3} lists some examples:
\begin{itemize}[noitemsep]
    \item \textbf{Type 1:} A sentence that does not contain in-line citations and is followed by a sentence of the same type. Such sentences are labeled `\textit{\emph{\textbf{0}}'.}
    \item \textbf{Type 2:} A Sentence that does not contain in-line citations but is followed by a sentence containing in-line citations. Such sentences are ignored and not included in our dataset as we cannot classify them as citation-worthy or not with complete certainty.
    \item \textbf{Type 3:} Sentences that contains in-line citations. Such sentences are ignored and not included in our dataset as removing citations from them may lead to the incorrect grammatical structure of the sentence.
    \item \textbf{Type 4:}  A Sentence that does not contain in-line citations and is followed by a sentence that is a citation in itself. Such sentences are labeled `\textit{\emph{\textbf{1}}'.}
\end{itemize}

\subsection{Dataset Profiling}

Our final dataset contains 178 million sentences. We publish three versions of the dataset on huggingface: small, medium, and large (original). Table \ref{tab:datastat} presents complete dataset statistics and Table \ref{tab:tablealldata} describes the dataset sizes for different versions.

\begin{table}[hbt!]
    \centering
    \begin{tabular}{l|r}
        Metric & \#  \\
        \hline
        &  \\
        Total Sentences & 178,459,203\\
        Total Files & 5,548,618\\ 
        Train Sentences & 142,588,927\\ 
        Train Files & 4,434,179 \\ 
        Dev Sentences &  17,934,940 \\ 
        Dev Files & 557,541 \\ 
        Test Sentences & 17,935,336 \\ 
        Test Files & 556,898 \\ 
        Total citation-worthy  & \multirow{2}{*}{10,487,177}\\ 
        sentences & \\
        Total non-citation-worthy & \multirow{2}{*}{167,972,026} \\ 
         Sentences & \\
        Avg character length of  & \multirow{2}{*}{171.61} \\ 
        citation-worthy Sentences & \\
        Avg character length of  & \multirow{2}{*}{172.93} \\ 
        non-citation-worthy Sentences & \\
        Number of Sentences  & \multirow{2}{*}{32.16} \\ 
        extracted per document & \\
    \end{tabular}
    \caption{\label{tab:datastat} Different Statistics on the CiteCaseLAW dataset.}
\end{table}

\begin{table}[!htbp]
    \centering
    \resizebox{\columnwidth}{!}{%
    \begin{tabular}{l|c|c}
        Dataset  & Total Sentence & Citation-Worthy  \\
        Version & Count & Sentences \\
        \hline
         & & \\
        Large & 178,459,203 & 10,487,177 ($\sim$5.87\%)  \\
        Medium & 10,000,000 & 586,999 ($\sim$5.869\%)  \\
        Small & 1,000,000 & 58,909 ($\sim$5.89\%)  \\
    \end{tabular}
    }
    \caption{Statistics of the different versions of the CiteCaseLAW dataset.}
    \label{tab:tablealldata}
\end{table}

Manual validation of 1000 random sentences to check the assigned labels was carried out based on two parameters -- accuracy of citation splitting mechanism and citation detection. Only Eleven sentences were incorrectly split, either in multiple sentences or got split partially. Three of these were due to OCR inconsistencies, leading to an accuracy of 98.9\%. Excluding these inconsistencies, the accuracy of splitting becomes 99.2\%. Labels assigned to each sentence were also correct.

\section{Methodology}

In this section, we experimented with different models trained on our dataset to establish the baselines for the task of citation-worthiness detection (\textit{RQ2}). For this assessment, we used our small split of the dataset of 1M entries. The split contains citation-worthy sentences equivalent to the big curated dataset sampled over all jurisdictions. A thorough hyperparameter search is done and is mentioned in the appendix section \ref{hyperparameters}. The models used in our research are as follows:-
\begin{itemize}
\item\textbf{Logistic Regression} This is a simple baseline with TF-IDF as input features.
\item\textbf{CRNN} A Convolutional Recurrent Neural Network in similar architecture as \citeauthor{farber2018cite} with little modifications.
\item\textbf{Transformer} A transformer \cite{vaswani2017attention} is trained from scratch. 
\item\textbf{Longformer} A transformer-based model designed to handle longer sequences and uses a sparse attention mechanism introduced by \cite{beltagy2020longformer}.
\item\textbf{BERT} A popular model with a strong reputation due to its performance on various tasks, we selected BERT \cite{devlin2018bert} for our classification task.
\item\textbf{LEGAL-BERT} A member of the family of BERT model pre-trained on large legal corpora spanning across different countries. Developed by \citeauthor{chalkidis2020legal} for legal domain, and as our domain matches it, the model is best suited.
\item\textbf{LEGAL-BERT+PU} The intention is to reduce subjectivity when writing a citation-worthy sentence. Figure \ref{fig:pulearn} shows the working diagram of positive unlabeled (PU) learning.   Previously, PU learning has shown promising results in rumor detection on Twitter and citation needed detection in Wikipedia \cite{wright2020claim}. The basis of PU learning is to suppose that positive, i.e., cite-worthy data is labeled, and non-citation-worthy data is unlabelled. A classifier is trained on the positive and unlabeled data to estimate that a given sample is labeled. Using the classifier, we estimate whether a sample is positive, given its unlabelled. We then combine positive samples with one copy of unlabelled samples marked as positive and the other as negative. The unlabelled samples are then weighed by the first classifier's estimate of the probability of the sample being positive. Finally, a classification model is trained on the task of citation-worthiness.

\begin{figure*}
    \centering
    \includegraphics[width=11.5cm,height = 5.5cm]{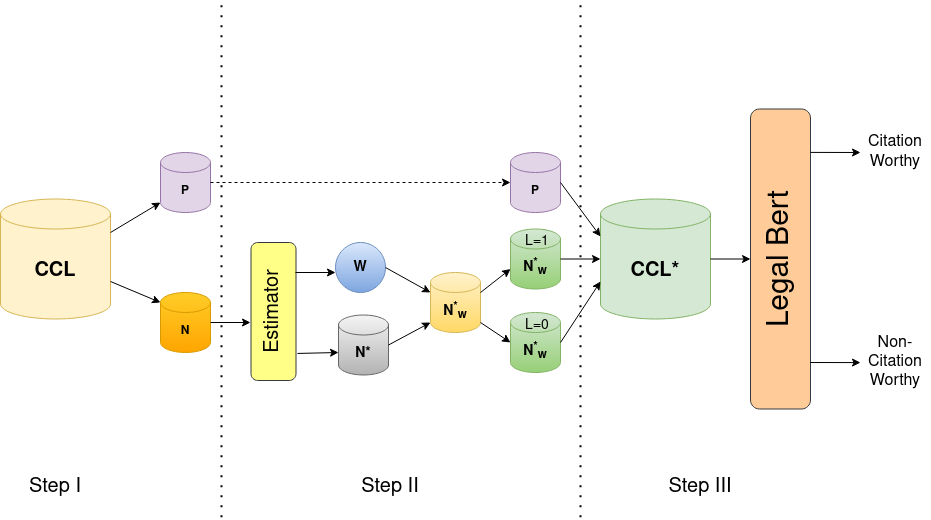}
    \caption{Here CCL is the CiteCaseLaw datset. It is divided into positive (P) and negative (N) samples while positive being unperturbed. Negative samples are passed through an estimator to get weights of samples of which model think it can have a label of citation worthy. Each sample is weighed using the weights produced $(N^{*}_{w[L=1]})$ and we duplicate these samples to have a label of citation unworthy $(N^{*}_{w[L=0]})$. These are concatenated to form $CCL^{*}$ and finally we predict the labels. }
    \label{fig:pulearn}
\end{figure*}
\end{itemize}

We examined the model's performance on other legal tasks using the datasets described below.

\textbf{UNFAIR-ToS} \cite{lippi2019claudette} There are 50 Terms of Service (ToS) from online platforms, including YouTube, eBay, Facebook, and others, in the UNFAIR-ToS dataset. Eight categories of unfair contractual terms or phrases (sentences) that may violate user rights under EU consumer legislation have been annotated in the dataset at the sentence level. A sentence is the model's input, and its output is a set of unfair kinds (if any). 

\textbf{LEDGAR} (Labeled EDGAR) \cite{tuggener2020ledgar}, a dataset for contract provision (paragraph) classification, was introduced in 2020. The terms of the contracts were gleaned from US Securities and Exchange Commission (SEC) filings, which are accessible to the general public through EDGAR10 (Electronic Data Gathering, Analysis, and Retrieval system). The original dataset contains roughly 850k contract clauses that are divided into 12.5k categories. This is a single-label multi-class classification job where each label reflects the single principal topic (theme) of the related contract clause.

We need to analyze the performance of the model after fine-tuning on citation-worthiness task for two important reasons. First, to demonstrate that these fine-tuned models did not perform poorly on already established baselines. Second, our PU learning model is based on the same LEGALBERT, so it should perform on par with vanilla LEGALBERT on baselines.

\section{Evaluation}

Table \ref{tab:baselines} shows the classification performance of the models. The pre-trained transformer models outperformed logistic regression and other deep-learning models. Introducing domain knowledge to the pre-trained models enhanced their performance. Adding PU learning increases the fraction of relevant instances retrieved, thereby making the model robust to citation-worthiness detection tasks after removing subjectivity. This answers our \textit{RQ3}. It showed $\sim$1\% improvement over LEGAL-BERT.

Table \ref{tab:transfer_dataset} shows the micro and macro F1 scores based on Transfer Learning on the datasets for other legal text classification tasks. LEGALBERT was established as baselines on UNFAIR-Tos  and LEDGAR datasets.
It is observed that fine-tuning the language model on our data suffices to enhance the performance. It is consistent with prior research indicating that improving language model fine-tuning on in-domain data results in improved end-task fine-tuning \cite{gururangan2020don}. This answers our \textit{RQ4} that model will not degrade after fine-tuning and will be at least comparable to the baselines.

\begin{table}[!ht]
    \centering
    \begin{tabular}{l|c|c|c}
        Model & P & R & F1 \\
        \hline
        & & & \\
        
        Logistic Regression & 77.85 & 75.77 & 76.79 \\
        CRNN & 76.54 & 74.72 & 74.93 \\
        Transformer & 72.42 & 84.25 & 77.89 \\
        Longformer & 87.10 & 86.02 & 86.56 \\
        BERT & \textbf{87.73} & 86.56 & 87.14 \\
        LEGAL-BERT & 87.64 & \textbf{87.2} & \textbf{87.42} \\
        \hline
        LEGAL-BERT + PU & 84.17 & \textbf{92.86} & \textbf{88.30} 
        \\

    \end{tabular}
    \caption{Classification results on the dataset of different models. Legal-BERT performed better than all mentioned models, taking a step further we applied PU learning over it to remove subjectivity answering our \textit{RQ4}.}
    \label{tab:baselines}
\end{table}

\begin{table*}[!h]
    \centering
    \begin{tabular}{l|c|c|c|c}
\hline
\multicolumn{1}{c}{\multirow{2}{*}{\textbf{Model}}} & \multicolumn{2}{l}{\textbf{UNFAIR TOS Dataset}}  & \multicolumn{2}{l}{\textbf{LEDGAR Dataset}}      \\ 
\multicolumn{1}{c}{}                               & $\mu-F1$ & m - F1 & $\mu-F1$ & m - F1 \\ 
\hline
& & & &\\
LEGAL-BERT                                          & 96.0                           & 83.0            & 88.2                           & 82.5            \\
LEGAL-BERT - CiteCaseLAW                            & \textbf{96.2}                  & \textbf{84.2}   & 88.2                           & \textbf{83.0}   \\
LEGAL-BERT+PU - CiteCaseLAW                         & 96.1                           & 83.5            & \textbf{88.4}                  & 82.7           
\end{tabular}%
    \caption{Results of F1 score based on Transfer Learning on Legal datasets. Comparable performance showed that fine-tuning with cite-worthiness data did not lead to any performance degrade}
    \label{tab:transfer_dataset}
\end{table*} 

 \section{Conclusion}
In this research, we constructed a large and novel dataset for the citation-worthiness task in the American legal domain. We analyzed various models and discovered that the domain-specific pre-trained language models generally outperform other models. Also we tried to remove subjectivity from the model. The legal community could use these models to identify citation-worthy sentences while drafting judgments. CiteCaseLAW is a valuable test platform for transfer-learning setup by demonstrating the models' suitability for downstream natural language understanding tasks.
We anticipate that the research community addressing problems in the field of legal language processing will find this data and associated fine-tuned models beneficial.

\section{Limitations}
We experimented upon a small split of our dataset, which took ~36 hours on each epoch. However, we created the dataset utilizing the complete CAP corpus and made it publically available for people in the legal domain to utilize in different tasks. The extension of our research to legal citation recommendation task can also be addressed by considering the metadata containing citation links. 


\bibliography{custom}
\bibliographystyle{acl_natbib}

\clearpage

\appendix


\section{Citation Detection}
We follow a multi-step methodology to detect all the citations present in the legal text.\footnote{Code for citation detection has been submitted in the supplementary material.} \\
Regex to detect the boundary of versus type cases:

\verb^([A-Z][A-Za-z-’]+|[A-Z]\.)(\s([A-Z]\.^
\verb^|of|and|&)|(?:\s[A-Z][A-Za-z-’]*))*^

The above regex was used both before and after the occurrence of `v.' in order to identify both the parties involved in the case. \\
We replace all the citations detected with a placeholder\\ \emph{[CITATION\textunderscore SPAN]}. 

\section{Sentence Boundary Detection} \label{appendixsbd}

Table \ref{tab:segtokexamples} gives toy examples after usage of different sentence boundary detectors on our dataset. Following are few examples of citations from the corpus:

\begin{enumerate}
\item \emph{168 Pa. Superior Ct. 351, 77 A. 2d 706}
\item \emph{State v. Camerlin, 117 R.I. 61, 362 A.2d 759 (1976)}
\item \emph{Interstate Coal Co. v. Trivett, 155 Ky. 825, 160 S. W. 728}
\end{enumerate}

\begin{table*}[h]
    \begin{tabular}{p{1.5cm}@{}p{13cm}@{}}
        \hline
        \vspace{1pt}
 \textbf{Original}&\vspace{1pt}\textcolor{Aquamarine}{The copy of the hospital record, being a photostat, was admissible under Code (1427), Art. 4335, sec. 3459, and was produced by mr. Alex, who was in charge at the time.} \textcolor{DodgerBlue3}{Copies of the statement given by Tyler to the police and the police report, were likewise properly put in evidence through the investigating officer.}
 \vspace{1pt}
        \\
        \hline
        \vspace{1pt}
    \textbf{simple period split (.)}&\vspace{1pt}\textcolor{Aquamarine}{The copy of the hospital record, being a photostat, was admissible under Code (1427), Art}
    
    \textcolor{Aquamarine}{ 4335, sec}
    
    \textcolor{Aquamarine}{ 3459, and was produced by mr}
    
    \textcolor{Aquamarine}{ Alex, who was in charge at the time}
    
    \textcolor{DodgerBlue3}{Copies of the statement given by Tyler to the police and the police report, were likewise properly put in evidence through the investigating officer.}
    \\
        \hline
        \vspace{1pt}
    \textbf{SegTok}&\vspace{1pt}\textcolor{Aquamarine}{The copy of the hospital record, being a photostat, was admissible under Code (1427), Art.}
    
    \textcolor{Aquamarine}{4335, sec.}
    
    \textcolor{Aquamarine}{3459, and was produced by mr.}
    
    \textcolor{Aquamarine}{Alex, who was in charge at the time.}
    
    \textcolor{DodgerBlue3}{Copies of the statement given by Tyler to the police and the police report, were likewise properly put in evidence through the investigating officer.}
    \\
    \hline
    \vspace{1pt}
    \textbf{Spacy blackstone}&\vspace{1pt}\textcolor{Aquamarine}{The copy of the hospital record, being a photostat, was admissible under Code (1427), Art.}
     
     \textcolor{Aquamarine}{4335, sec.} 
     
     \textcolor{Aquamarine}{3459, and was produced by mr.} 
     
     \textcolor{Aquamarine}{Alex, who was in charge at the time.} 
     
     \textcolor{DodgerBlue3}{Copies of the statement given by Tyler to the police and the police report, were likewise properly put in evidence through the investigating officer.}
        \\
        \hline
        \vspace{1pt}
        \textbf{pySBD}&\vspace{1pt}\textcolor{Aquamarine}{The copy of the hospital record, , being a photostat, was admissible under Code (1427), Article 4335, section 3459, and was produced by mr Alex, who was in charge at the time.}
    
    \textcolor{DodgerBlue3}{Copies of the statement given by Tyler to the police and the police report, were likewise properly put in evidence through the investigating officer.}
    \\
    \hline
    \end{tabular}
    \caption{Toy examples showing performance of different segmenters for unstructured legal text. Due to presence of many abbreviated/short terms followed by period (.) in legal text, it makes difficult for segmenters to decide the point of segmentation. Apart from this a simple example would be of reporters in citations i.e. \textit{\textbf{in lorem vs ipsum co. ltd. 123 S. Ct. 456, 789.} These have similar outputs which shows they could not correctly split the sentences. As it is evident in the above examples, a single sentence was broken down into four different sentences because of the presence of a period after the acronyms `art', `sec', and `mr'. The sentence splitter failed to recognize this and produced incorrect results.} }
    \label{tab:segtokexamples}
\end{table*}

List of acronyms/shortenings which caused incorrect sentence splitting. These were identified and replaced with their version which didn't contain a full stop.
\begin{itemize}[noitemsep]
\begin{minipage}{0.30\linewidth}
\item Inc.                                           
\item Co.                       
\item Ltd.   
\item No.    
\item Vol.    
\item Corp.       
\item Viz.    
\item Mfg.                                           
\item Dist.                                     
\item Commn.                                     
\item Sec.                                  
\item Pet.                           
\item Com.                           
\item Eq.                           
\item Doc. 
\end{minipage}
\begin{minipage}{0.30\linewidth}
\item Ins.                                       
\item Ex.                                            
\item Cf.                                        
\item Civ.
\item a.m.
\item p.m.
\item e.g.
\item Pvt.                                
\item Ms. 
\item Mr.  
\item Jr.          
\item Sr.
\item Dr.
\item Al.               
\item A.
\end{minipage}
\begin{minipage}{0.30\linewidth}
\item Q.                                             
\item Cont.                                      
\item Aff.                                           
\item Cert.                                           
\item Art.                                           
\item Bros.                                           
\item Ref. 
\item Mrs.                                            
\item Ed.                                      
\item Nom.                      
\item Ch.                    
\item Eq.   
\item D.C.
\item i.e.
\end{minipage}
\end{itemize}
Apart from these, several instances of multiple consecutive punctuation marks were also fixed.\\
Once the individual sentences were split, the following regex was used to detect if a sentence was a citation in itself:
\verb|^(See)?(\s)?(eg)?(\s)?(\[CITATION\_SPAN\]|
\verb|\s?)+$|

\section{Data Visualization}
Our data was divided into 61 different jurisdictions according to data present in https://cite.case.law/. For visualizing the data, we used legal cases from the following jurisdictions:
\begin{itemize}[noitemsep]
    \item Louisiana (la)
    \item Illinois (ill)
    \item Arkansas (ark)
    \item Massachusetts (mass)
    \item Wisconsin (wis)
\end{itemize}
For visualizing the data, TSNE plots were made. The hyper parameter perplexity was set at 2000 for the plot corresponding to state wise division and it was set at 200 for the plot corresponding to the century wise division. \\

Further details of the TSNE plots can be found at: \url{https://scikit-learn.org/stable/modules/generated/sklearn.manifold.TSNE.html}.

\section{Experimental Setup}
\subsection{Infrastructure}
A system with 48 cores with muliple GPUs and $\sim$ 500GB (not even 40\% utilised) RAM was used in the experimentation. All training were done using GeForce RTX 3090 with a memory of 24,268 MB (~24GB).

\subsection{Hyperparameter Tuning} \label{hyperparameters}

Here, except for logistic regression, we used Ax search. It can be found here: \url{https://docs.ray.io/en/latest/tune/api_docs/suggestion.html#ax-tune-suggest-ax-axsearch}. For logistic we used sklearn's built in \textit{RandomizedSearchCV}. See table \ref{tab:hyperparameters}\\

\subsubsection{Logistic Regression} For Logistic regression, a search space of \textit{spacy.stats.uniform(loc=0, scale=4)} was taken with L1 and L2 regularization. Selected parameters were C: 0.1151395399 and regularization: L2.\\

Further documentation for \textit{uniform} function can be found at: \url{https://docs.scipy.org/doc/scipy/reference/generated/scipy.stats.uniform.html}.\\

\subsubsection{CRNN} For CRNNs the search space for learning rate, epoch and batch size is [1e-3, 1e-2], [3, 15], \{4,8,32,128\} respectively. The selected parameters as learning rate: 0.00523737; epochs: 3 and batch size: 32.\\

\subsubsection{Tranformer} For transformers a search space for learning rate, weight decay, warmup steps, epochs, feed forward layers, number heads, epochs and dropouts  are: [1e-7, 1e-3], \{0.0, 0.0001, 0.001, 0.01, 0.1\},	\{0, 100, 200, 300, 400, 500, 1000, 1500, 2000, 2500, 5000\},	[3, 35], \{128,256,512,1024,2048\}, \{1,2,3,4,5,6,10,12\}, \{0.0,0.1,0.2,0.3,0.4,0.5\} respectively. The selected values in the mentioned order are: 0.000174364, 0.1, 14, 128, 5, 4, 0.5.\\

\subsubsection{Xformer} The learning rate  was tuned in the range [1e-7, 1e-4], while with BERT, the rate is in the range [1e-8, 1e-5]. We used a triangular learning rate. Search space for Weight decay, Warmup steps and epochs are \{0.0, 0.0001, 0.001, 0.01, 0.1\}, \{0, 100, 200, 300, 400, 500, 1000, 1500, 2000, 2500, 5000 \}, [3, 15] respectively. The batch size was taken as 4. The selected parameters for the models are listed in Table \ref{tab:hyperparameters}.



\begin{table}[!h]
    \centering
    \begin{tabular}{c|c|c|c|c}
    \multirow{2}{*}{Model} & \multirow{2}{*}{LR (x$10^{-6}$)} & \multirow{2}{*}{Decay} & Warmup & \multirow{2}{*}{Epochs}\\
    & & & Steps &\\
    \hline 
    & & & & \\
    BERT & 9.8276 & 0.01 & 1000 & 8 \\
    LEGALBERT & 9.6984 & 0.1 & 300 & 11 \\
    LEGALBERT & \multirow{2}{*}{6.6823} & \multirow{2}{*}{0.0} & \multirow{2}{*}{400} & \multirow{2}{*}{3} \\
    + PU & & & & \\
    Longformer & 7.2936 & 0.1 & 2000 & 9 \\
    \end{tabular}
    \caption{Selected hyperparameters for different models}
    \label{tab:hyperparameters}
\end{table}


\section{Metrices}

We used sklearn's \textit{precision\_recall\_fscore\_support} for the following metrices\\

$Precision = \frac{TP}{TP+FP}$\\

$Recall = \frac{TP}{TP+FN}$\\

$F1 = \frac{2*Precision*Recall}{Precision+Recall} = \frac{2*TP}{2*TP+FP+FN}$\\

Further documentation can be found at: \url{https://scikit-learn.org/stable/modules/generated/sklearn.metrics.precision_recall_fscore_support.html}.\\

Here $T$ stands for True, $F$ for False, $P$ for positives and $N$ for negatives. hence $TP$ stands for true positives and so on.

\subsubsection{Macro F1} Macro F1 is the average of F1 scores of all the classes.
\subsubsection{Micro F1} Micro F1 is the weighted sum of F1 scores of all the classes where weights are the class distribution in the dataset.


\section{Dataset and statistics} \label{appendixdatastats}
We present our final dataset in jsonl format where each sentence is an object having the following parameters: 

\subsection{Meta-data} \label{appendixmeta}
\begin{itemize}
\item \textbf{File Name:} The case file to which the sentence belongs. 
\item \textbf{Sentence Number:} The sentence number as present in the document.  
\item \textbf{Sentence:} The naturally occurring sentence in the text (after preprocessing/removing citation span.)
\item \textbf{Label:} Integer value of `\textit{\emph{\textbf{0}}}' or `\textit{\emph{\textbf{1}}}'. `\textit{\emph{\textbf{0}}}' represents that the sentence is not citation-worthy whereas `\textit{\emph{\textbf{1}}}' represents that the sentence is citation-worthy.
\end{itemize}




\subsection{Data Pre-processing} \label{sectionG.3}

We handled acronyms, like `article' instead of `art.', `section' instead of `sec.', `number' instead of `no.' and so on. Some of these acronyms/shortenings are commonly used for e.g. `no.', `i.e.', `ms.' whereas some were legal jargon like `cf.', `D.C.', `Inc.'. We also identified different bodies, laws, and sections whose names contained `.' and were commonly referred to in the case laws of American legal corpus. After including the aforementioned steps, we use the pySBD module to prepare the final dataset. An example is given in Table \ref{tab:processedSentencesExamples}. 

In all the dataset versions, we have followed nearly an 80:10:10 split for train, validation, and test sets respectively. The split is document level which means that all the sentences belonging to the same document will only be present in one of the train, validation or test splits.

\begin{table*}
    \vspace{3pt}
    \begin{tabular}{p{0.5cm}p{\textwidth}}
    (1) & \textbf{[ORIGINAL]} On appeal to this Court, we held that the railroad had acquired by condemnation proceedings a base or conditional fee, terminable on the cesser of the use for railroad purposes. Lacy v. East Broad Top Railroad and Coal Co., 168 Pa. Superior Ct. 351, 77 A. 2d 706. \vspace{2pt}\\
     & \textbf{[PROCESSED]} On appeal to this Court, we held that the railroad had acquired by condemnation proceedings a base or conditional fee, terminable on the cesser of the use for railroad purposes.
     \vspace{3pt}\\
     \hline
     \vspace{1pt}
    (2) & \vspace{1pt}\textbf{[ORIGINAL]} In Tanorio v. Superior Court, 1 N.Mar.I. 4, we determined under what conditions a writ of mandamus may issue.\vspace{2pt}\\ 
     & \textbf{[PROCESSED]} \textit{\emph{\textbf{Ignored}}.} (citation present at the start of the sentence.)
     \vspace{3pt}
    \end{tabular}
    \vspace{3pt}
    \caption{Excerpt from training samples in CiteCaseLAW. The first example belongs to Type 4 sentences whereas the second example belongs to Type 3 sentences. }
    \label{tab:processedSentencesExamples}
\end{table*}

\end{document}